\documentclass{article} 

\usepackage{iclr2027_conference,times}

\usepackage{amsmath,amsfonts,bm}

\def\eqref#1{equation~\ref{#1}}

\def\1{\bm{1}}

\DeclareMathAlphabet{\mathsfit}{\encodingdefault}{\sfdefault}{m}{sl}
\SetMathAlphabet{\mathsfit}{bold}{\encodingdefault}{\sfdefault}{bx}{n}

\usepackage{url}
\usepackage{amsmath}
\usepackage{amssymb}
\usepackage{booktabs}
\usepackage{multirow}
\usepackage{graphicx}
\usepackage{float}
\usepackage{hyperref}
\hypersetup{hidelinks}
\iclrfinalcopy 

\title{PortraitAes: Intent-Conditioned Structured Portrait Aesthetic Assessment}

\makeatletter
\g@addto@macro\@thanks{%
  \begingroup
  \footnotetext{\noindent\raggedright
    $^{\ddagger}$These authors contributed equally. \quad
    $^{*}$Co-corresponding authors. \quad
    $^{\dagger}$Project lead.}%
  \endgroup
}
\makeatother

\author{
  \begin{tabular}{@{}c@{}}
    \textbf{Junzhou Xie}$^{1,2,\ddagger}$ \hskip 0.35em
    \textbf{Haozhong Xiong}$^{2,\ddagger}$ \hskip 0.35em
    \textbf{Xunyun Tian}$^{2}$ \hskip 0.35em
    \textbf{Kaile Du}$^{1}$ \hskip 0.35em
    \textbf{Tianchen Yu}$^{2}$ \hskip 0.35em
    \textbf{Qiang Li}$^{2}$ \\[0.5ex]
    \textbf{Wei Liu}$^{2}$ \hskip 0.35em
    \textbf{Jiaming Liu}$^{2}$ \hskip 0.35em
    \textbf{Ruihua Huang}$^{2}$ \hskip 0.35em
    \textbf{Yang Shi}$^{2,*,\dagger}$ \hskip 0.35em
    \textbf{Guangcan Liu}$^{1,*}$ \\[1.5ex]
    \textmd{$^{1}$Southeast University \qquad $^{2}$Qwen Business Unit of Alibaba} \\[0.5ex]
    \textmd{\texttt{newsy0106@gmail.com}, \texttt{guangcanliu@seu.edu.cn}}
  \end{tabular}
}

\newcommand{\bench}{PortraitAes-Bench}
\newcommand{\method}{PortraitAes}

\begin{document}

\maketitle
\lhead{}
\renewcommand{\headrulewidth}{0pt}
\vspace{-0.5em}
\begin{abstract}
Portrait aesthetic assessment assigns comparable scores according to how effectively human-centered images fulfill their photographic intent. These scores support data filtering, candidate selection, and preference modeling in image-generation pipelines. Existing methods typically predict a single aesthetic score or use general-purpose MLLMs without conditioning on photographic intent. This omission matters because the same blur, pose, lighting, or framing choice may serve one photographic intent but undermine another. These models thus learn context-agnostic aesthetic priors and yield inconsistent, inaccurate, misleading judgments for portraits with distinct photographic objectives.
We introduce \bench, an 11K-scale benchmark that decomposes this task into intent-conditioned subjudgments. Expert-authored rubrics define nine photographic intents, six first-level dimensions, and 22 secondary criteria. They support a structured pipeline for intent routing, specialist assessment, verification, and score fusion. Following this structure, we train \method{} with multi-task supervision. We then improve score comparability through Gaussian score calibration and within-dimension cross-image ranking. On the standard benchmark, \method{} achieves a Pearson correlation of 0.924 and a Spearman rank correlation of 0.934. On the hard-case set, its Pearson correlation is 0.829 and its Spearman rank correlation is 0.795. Across both sets, \method{} outperforms the evaluated general-purpose MLLMs and specialized aesthetic baselines. Project page: https://portraitaes.github.io/.

\end{abstract}

\section{Introduction}


Large-scale image generation pipelines rely on automated aesthetic assessment for data filtering, candidate output selection, and preference or reward signal learning~\citep{schuhmann2022laion,kirstain2023pick,xu2023imagereward}. In this work, we investigate \emph{portrait aesthetic assessment}: given a human-centric image, the task aims to interpret how its visual composition supports photographic intent and encode such perceptual judgments into comparable scalar scores. Unlike facial attractiveness prediction, this task evaluates the holistic presentation of a subject, encompassing framing, pose, gesture, action, appearance, and contextual interplay with the surrounding scene.

Portraits serve functional, commercial, documentary, social, and artistic purposes. The same blur, lighting, framing, or pose may weaken an identity portrait yet support a narrative or artistic one. A useful assessor must therefore infer intent, diagnose strengths and failures in structure, composition, image quality, tone, theme and style, and emotion, ground each judgment in visible evidence, and preserve ordering across portraits. Portrait assessment is thus an intent-conditioned, structured evaluation task rather than single-score prediction.

Existing resources cover generic aesthetic ratings, perceptual quality, multimodal critique, and portrait-specific attributes~\citep{murray2012ava,kong2016photo,huang2024aesbench,zhou2024uniaa,huang2022aesthetic,sha2026portraitcraft}. Their labels seldom expose photographic intent, dimension-specific rules, visible evidence, and intent-conditioned fusion together. Single holistic scores are incapable of pinpointing specific flaws in an image, whereas naive scoring with MLLMs suffers from the absence of consistent photographic standards and criterion-aligned optimization targets. The missing supervision is not another global preference score, but an executable account of which evidence matters under a given intent and how the resulting judgments should be combined.

We introduce \bench, an 11K-scale benchmark that operationalizes expert-authored rubrics through a router--specialist--fusion workflow. A router infers one of nine photographic intents and its dimension weights; six specialists assess structure, composition, image quality, tone, theme and style, and emotion; intent-conditioned fusion produces the overall score. A high-capacity proprietary multimodal LLM executes these rubric-bound roles, followed by schema checks, Qwen3.5 Flash screening, and expert verification. Each of 1.6K images yields one routing record and six specialist records, producing 11K records with scores, subcriteria, semantic labels, visible-evidence rationales, and veto/cap decisions. This role-based construction mirrors the task structure instead of asking the annotation backend for an unconstrained holistic judgment.

Using the same seven-task structure, \method{} trains a shared Qwen3.5-VL-9B model~\citep{qwen359b2026} through multi-task supervised fine-tuning (SFT). Gaussian rewards then anchor predictions to rubric-derived scores, while cross-group Bayesian personalized ranking (BPR) enforces relative order between portraits evaluated by the same specialist. Semantic, format, and Kullback--Leibler (KL) terms retain the structured outputs. \method{} achieves overall PLCC/SRCC values of 0.924/0.934 on the standard benchmark and 0.829/0.795 on the hard-case set; adding cross-group BPR to Gaussian-only training improves both hard-case correlations by 0.040.

Our main contributions are summarized as follows:

\begin{itemize}
    \item \textbf{An intent-conditioned portrait aesthetic benchmark.} \bench{} turns expert-authored rubrics into 11K structured records for 1.6K portraits, covering nine photographic intents, six dimensions, and 22 secondary criteria through intent routing, specialist assessment, and score fusion.

    \item \textbf{A calibrated and rank-aligned portrait aesthetic model.} \method{} learns the same routing and specialist structure through multi-task SFT. Gaussian calibration aligns predicted scores with rubric-derived references, while same-dimension cross-group ranking improves relative ordering across portraits.

    \item \textbf{Empirical gains on standard and hard cases.} \method{} achieves the strongest correlations among the evaluated general-purpose and specialized baselines on both the standard benchmark and its range-restricted hard-case set. Component ablations isolate the effects of structured SFT, Gaussian calibration, and cross-group ranking.
\end{itemize}

\section{Related Work}

\subsection{General-purpose image aesthetic and quality assessment}
Portrait assessment builds on image aesthetics and perceptual quality assessment.
General aesthetic work spans population ratings and attributes, score distributions, visual backbones, native-resolution quality, and personalization~\citep{murray2012ava,kong2016photo,talebi2018nima,hosu2019effective,ke2021musiq,yang2022personalized}. In-the-wild IQA instead emphasizes authentic distortions and capture conditions~\citep{hosu2020koniq,fang2020perceptual,ying2020patches}. Multimodal benchmarks add critique, language-grounded judgments, and perceptual dimensions~\citep{vera2022understanding,huang2024aesbench,zhou2024uniaa,cao2026artimuse,cao2025unipercept,li2026qwen}. Together they provide broad supervision but rarely expose photographic intent. \bench{} links intent to portrait evidence and score fusion.

\subsection{Portrait-specific aesthetic and quality assessment}

Portrait-specific work adapts these objectives to human-centered images.
Prior studies cover spatial and lighting features, beauty prediction, AVA-derived protocols, MPAD/MPAAD, smartphone quality, and quality question answering~\citep{khan2012evaluating,redi2015beauty,kairanbay2016aesthetic,huang2022aesthetic,chahine2023image,wu2026q}. PortraitCraft studies portrait composition through global scores, attributes, visual question answering, and structured descriptions~\citep{sha2026portraitcraft}. \bench{} instead evaluates full human-centered images, not facial attractiveness or composition alone. Its six dimensions connect body framing, pose, action, gesture, posture, and scene interaction to explicit photographic intent.

\subsection{Aesthetic preference learning and reward models}

Preference learning connects aesthetic assessment with reward modeling for image generation.
Aesthetic supervision uses absolute ratings or score distributions, while pairwise objectives learn preference~\citep{kong2016photo,talebi2018nima}. For text-to-image generation, Pick-a-Pic provides prompt-conditioned choices, while ImageReward and HPS v2 learn preference scorers from comparisons~\citep{kirstain2023pick,xu2023imagereward,wu2023human}. MPS conditions evaluation on aesthetics, semantic alignment, detail quality, or overall preference~\citep{zhang2024learning}. These methods provide general rewards but do not organize portrait assessment by photographic intent and criterion-specific evidence. \method{} calibrates six rubric-defined dimension scores and performs ranking exclusively over examples sharing the same specialist dimension, while preserving their semantic labels and structured visual evidence.

\section{\bench}

\subsection{Benchmark overview}

\begin{figure}[t]
\centering
\includegraphics[width=\linewidth]{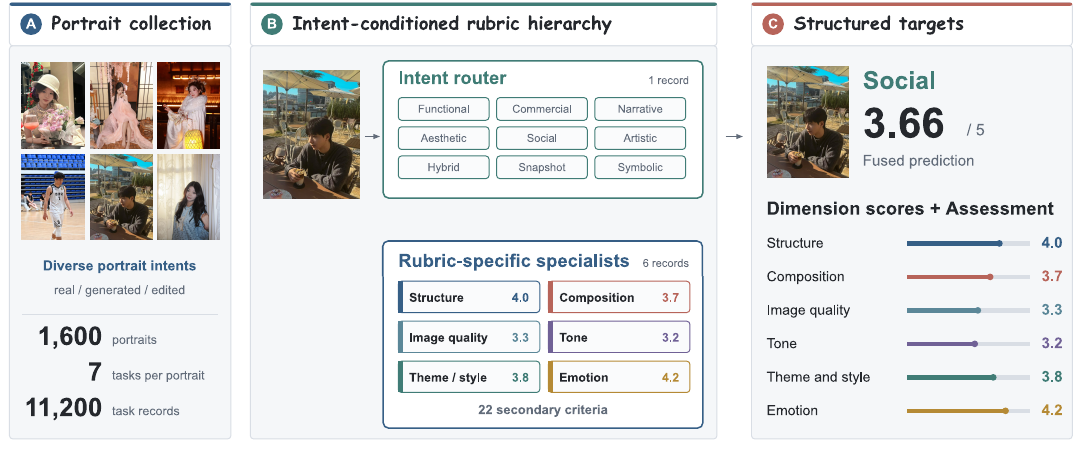}
\caption{Overview of \bench. Each portrait is routed to one of nine photographic intents and evaluated by six specialists spanning 22 secondary criteria, yielding structured routing, rating, and intent-conditioned fusion targets.}
\label{fig:bench-overview}
\end{figure}

\bench{} contains 1.6K portraits expanded into 11K task records, with one intent-routing record and six specialist records per image. It covers nine photographic intents, six dimensions (structure, composition, image quality, tone, theme and style, and emotion), and 22 secondary criteria. The records expose routing targets, dimension and subcriterion scores, semantic labels, visible-evidence rationales, veto/cap decisions, and an intent-conditioned overall score. \autoref{fig:bench-overview} summarizes this task structure; Appendix~\ref{app:resource-comparison} compares related resources.

The 11K records represent seven linked tasks for every portrait, rather than a collection of 11K independent portrait images with unrelated labels. The routing record supplies intent and fusion weights, while each specialist record preserves a dimension-specific judgment and its evidence. This organization supports separate evaluation of routing and specialist behavior as well as end-to-end evaluation after fusion.

\subsection{From photographic expertise to structured judgment}

Photographic expertise specifies what to inspect and how to judge it. Motion blur, for example, must be evaluated by whether the subject remains recognizable, whether the blur supports the depicted action, and whether that choice fits a functional, narrative, or artistic intent. \bench{} encodes this reasoning through intent categories, specialist criteria, score anchors, and decisions.

The design separates intent routing, specialist assessment, and score fusion. Routing establishes the photographic context and selects importance weights; specialists evaluate evidence under dimension-specific rubrics; fusion combines their judgments according to intent. This separates how well a criterion is realized from how much it matters to the portrait's function. A portrait can therefore be technically clean but emotionally weak, compositionally unconventional but artistically coherent, or noisy yet appropriate for a documentary or snapshot intent.

For image $x$, rubric $\mathcal{R}$ specifies the task boundary, labels, scoring criteria, output schema, and veto/cap rules. The router predicts intent $c$, its six-dimensional weight vector $w^{(c)}$, and visible routing evidence $e_{type}$:
\begin{equation}
(c, w^{(c)}, e_{type}) = f_{type}(x; \mathcal{R}_{type}).
\end{equation}
Appendix Table~\ref{tab:intent-weights} reports the weight matrix. Let $\mathcal{D}_{spec}$ denote the six specialist dimensions. For each $d\in\mathcal{D}_{spec}$, rubric $\mathcal{R}_d$ defines its labels, subcriteria, scoring rules, and output fields. Given the image and router context, the specialist predicts
\begin{equation}
z_d=(s_d, a_d, q_d, e_d, v_d)=f_d(x,c;\mathcal{R}_d),
\end{equation}
where $s_d\in[1,5]$ is the dimension score, $a_d$ denotes semantic labels, $q_d$ the subcriterion scores, $e_d$ the evidence-grounded rationale, and $v_d$ veto/cap decisions. For available specialists $D(x)$, the fused score is
\begin{equation}
S(x)=\frac{\sum_{d\in D(x)}w_d^{(c)}s_d(x)}{\sum_{d\in D(x)}w_d^{(c)}}.
\label{eq:fusion}
\end{equation}
Unavailable dimensions are removed from both sums rather than treated as zeros. Using reference or predicted routing weights also separates routing errors from specialist scoring errors.

\subsection{Executing expert rubrics for annotation}

The annotation pipeline turns these definitions into records through rubric-bound roles and quality control. Experts define the nine intents, six dimensions, 22 secondary criteria, score anchors, label spaces, task boundaries, and veto/cap rules. For each image, a high-capacity proprietary multimodal LLM serves as the shared annotation backend for the intent router and six specialist roles, executing each role under its corresponding rubric. The role definitions, schemas, and fusion rules are independent of the backend interface and support either open-weight deployments or proprietary APIs. Each role produces a separate structured record, connecting expert-defined criteria and exceptions to visible portrait evidence rather than relying on one unconstrained score. The schemas retain scores, labels, rationales, and rule decisions for supervision and audit.

Execution follows the dependency in the task definition. The router first produces the intent, weights, and routing evidence. Its context is then supplied to all six specialists, which inspect the same portrait through different criteria and return independently parseable records. Rule-based fusion is applied only after these records pass quality control, so the intermediate judgments remain available instead of being hidden inside the final score.

The rubrics specify contextual priorities and concrete observations. Functional and commercial intents prioritize recognizability and finish; narrative and symbolic intents emphasize theme and emotion; aesthetic and artistic intents allow stylization; social and snapshot intents emphasize presentation and affect. Hybrid intent balances these priorities. Structure covers subject prominence, completeness, and background harmony; composition covers framing, perspective, leading lines, and dynamism. Image quality examines sharpness, texture, noise, compression, and intent-sensitive tolerance. Tone relates light--dark organization and color treatment to brightness, purity, and harmony. Theme and style evaluates fit, consistency, and originality, while emotion connects affect and intensity to physiological cues, context, and body language.

\begin{figure}[t]
\centering
\includegraphics[width=\linewidth]{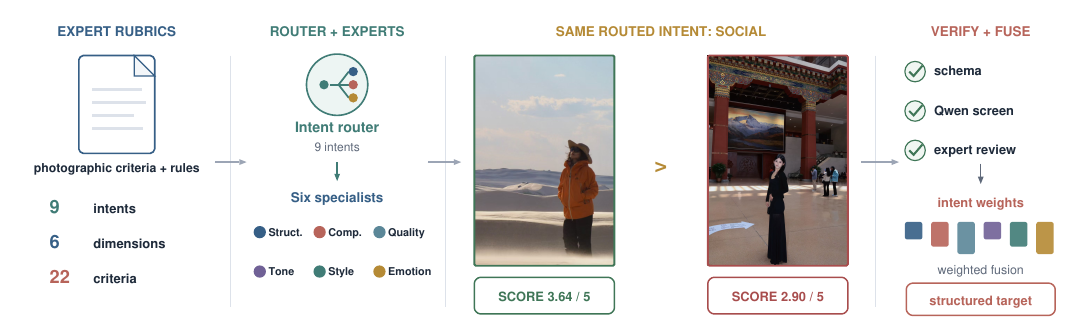}
\caption{Construction of \bench. Expert rubrics define an intent router and six specialist roles; schema checks, Qwen3.5 Flash screening, and expert review precede intent-weighted fusion.}
\label{fig:benchmark-construction}
\end{figure}

Before fusion, schema checks enforce the task contract and valid fields. Qwen3.5 Flash~\citep{qwen35flash2026} screens format, score ranges, label compatibility, and veto/cap consistency; experts verify rubric consistency and boundary cases. Specialist-level veto/cap decisions prevent hard dimension failures from being averaged away. \autoref{fig:benchmark-construction} summarizes the process.

\subsection{Artifact statistics}

After parsing, consistency screening, fusion checks, and expert verification, the artifact contains 1.6K images and 11K task records. Each image contributes one routing and six specialist records, supporting image-level and task-specific evaluation of routing, dimension-wise regression, semantic recognition, evidence-grounded explanation, and intent-conditioned scoring.

\begin{figure}[t]
\begin{center}
\includegraphics[width=\linewidth]{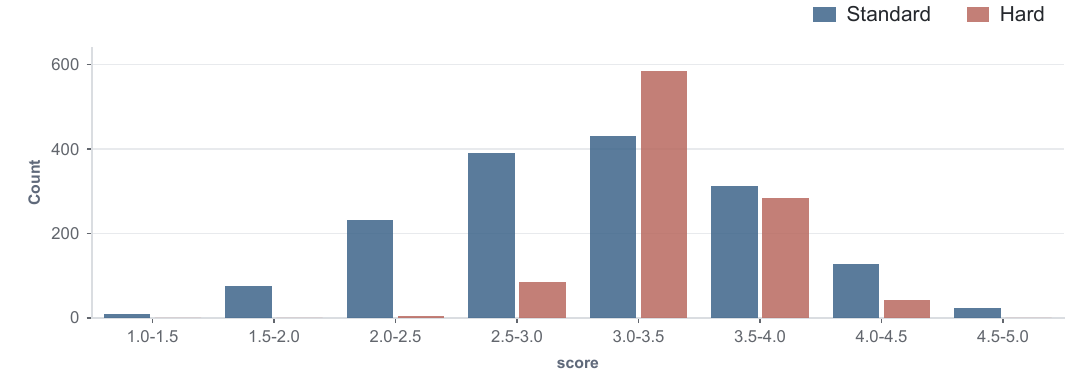}
\end{center}
\caption{Intent-conditioned overall-score distributions. The standard benchmark spans a broad range, whereas the hard-case set concentrates visually similar portraits for fine-grained ranking.}
\label{fig:overall-score-distributions}
\end{figure}

Appendix~\ref{app:diagnostics} summarizes the specialist-score distributions (\autoref{fig:dimension-score-distributions}). Emotion and structure concentrate higher, tone is lower and broader, and the remaining dimensions occupy the middle range, providing varied cases for calibration. The overall distributions contrast the broad standard benchmark with the deliberately range-restricted hard-case set (\autoref{fig:overall-score-distributions}).

\section{\method{}: Intent-Conditioned Portrait Aesthetic Modeling}

\begin{figure}[htbp]
\centering
\includegraphics[width=0.86\linewidth]{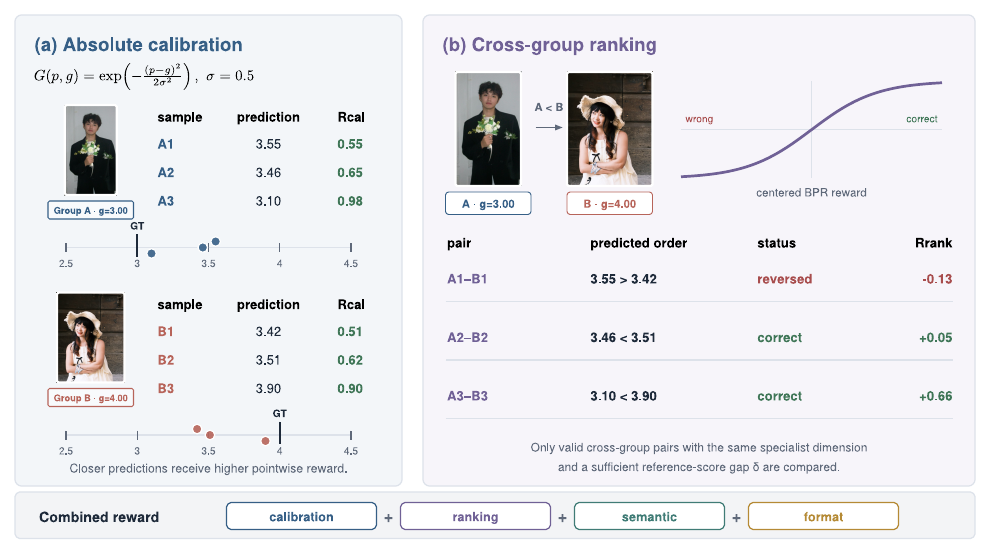}
\caption{Illustration of the numerical rewards. Gaussian calibration anchors predictions to rubric-derived scores; same-dimension cross-group ranking penalizes ordinal inversions. Images and rollout values illustrate the mechanism rather than report an evaluation case.}

\label{fig:grpo-reward}
\end{figure}

A portrait aesthetic model must explain its scores and compare images consistently. Holistic supervision leaves intent and individual criteria implicit; structured supervision exposes them, but independent predictions still do not constrain cross-image order and can rank a weaker portrait above a stronger one under the same criterion. Multi-task SFT first learns intent and specialist judgments under shared rubrics. Reinforcement learning (RL) then combines Gaussian calibration with same-dimension cross-group ranking, while semantic and format rewards preserve structured outputs (Figure~\ref{fig:grpo-reward}).

\subsection{Learning structured aesthetic priors}

Each training image yields one intent-routing target and six specialist targets, with a task-specific rubric and output schema. We construct the instruction--response examples using the same annotation protocol as \bench{}. The SFT corpus contains 10,286 images and approximately 72,000 examples across seven tasks, with 10,286 examples for each task: intent routing, theme and style, composition, structure, image quality, tone, and emotion. Qwen3.5 Flash normalizes output formats and corrects schema inconsistencies, followed by manual spot-checking for format compliance and annotation quality.

Let $a\in\mathcal{A}=\{r\}\cup\mathcal{D}$ index the routing task $r$ or one of the six specialist dimensions $d\in\mathcal{D}$, and let $q_a$ denote its task-specific instruction. Given an image $I$ and a structured target response $y$, we optimize the autoregressive objective
\begin{equation}
\mathcal{L}_{\mathrm{SFT}}
=-\mathbb{E}_{(I,q_a,y)\sim\mathcal{D}_{\mathrm{SFT}}}
\left[\sum_{t=1}^{|y|}\log\pi_\theta(y_t\mid I,q_a,y_{<t})\right].
\label{eq:sft-objective}
\end{equation}
Task conditioning lets one policy learn heterogeneous portrait rubrics over a shared visual representation without collapsing their label spaces or schemas. The same model can therefore execute one specialist in isolation or the complete router--specialist workflow. We apply low-rank adaptation (LoRA) to the Qwen3.5-VL-9B language backbone and freeze its visual encoder and aligner, concentrating adaptation on rubric interpretation, structured prediction, and score generation.

\subsection{Rank-aligned aesthetic optimization}

Token-level SFT learns rubric semantics and response structure but neither weights rating errors by numerical distance nor compares image queries. Predicting 1 for a reference of 5 is numerically worse than predicting 4, yet cross-entropy depends on target-token probability. Scores must therefore agree with their rubric-derived references and preserve same-dimension ordering. Small pointwise errors can reverse nearby scores, while predictions concentrated near the scale midpoint may have modest error but weak discrimination. We combine calibration with comparisons across prompt groups to address both problems.

\subsection{Intent-conditioned reward design}

The completion reward combines calibration, ranking, semantic consistency, and structural validity:
\begin{equation}
R_i=\lambda_{\mathrm{cal}}R_{\mathrm{cal},i}
+\lambda_{\mathrm{rank}}R_{\mathrm{rank},i}
+\lambda_{\mathrm{sem}}R_{\mathrm{sem},i}
+\lambda_{\mathrm{fmt}}R_{\mathrm{fmt},i},
\qquad \lambda_{\bullet}\ge 0.
\label{eq:reward}
\end{equation}
For specialist tasks, $R_{\mathrm{cal}}$ maps primary-score and subcriterion errors to Gaussian soft rewards~\citep{cao2025unipercept}. $R_{\mathrm{rank}}$ supplies the missing comparison signal: each rollout is compared only with valid cross-group examples from the same specialist dimension, using a BPR-style reward~\citep{rendle2012bpr} when the reference gap defines a reliable order. This constraint prevents comparisons between incompatible rubrics such as structure and emotion.

\textbf{Absolute calibration ($R_{\mathrm{cal}}$).}
Inspired by Gaussian soft rewards for visual rating~\citep{cao2025unipercept}, we map prediction error to a smooth calibration signal. For a scalar prediction $p$ and its rubric-derived reference $g$, we define
\begin{equation}
G(p,g)=\exp\!\left[-\frac{(p-g)^2}{2\sigma^2}\right],
\label{eq:absolute}
\end{equation}
where $\sigma$ controls the tolerance to score deviation. For a specialist dimension $d$ with primary score $(p_d,g_d)$ and rubric subcriteria $\mathcal{S}_d$, the calibration reward is
\begin{equation}
R_{\mathrm{cal}}^{(d)}=\eta_d G(p_d,g_d)
+(1-\eta_d)\sum_{k\in\mathcal{S}_d}\omega_{d,k}G(p_{d,k},g_{d,k}),
\qquad \sum_{k\in\mathcal{S}_d}\omega_{d,k}=1.
\label{eq:hierarchical-calibration}
\end{equation}
A specialist response must calibrate both its overall judgment and the evidence-bearing sub-scores.

\textbf{Cross-group relative ranking ($R_{\mathrm{rank}}$).}
Calibration evaluates each image independently and cannot constrain order between images. We therefore pool predictions from different GRPO completion groups within an update, which generally correspond to different image prompts. Each rollout is compared with valid same-dimension rollouts from other groups, and its ranking reward aggregates the pairwise terms.

Let $\mathcal{G}_i$ denote the prompt group that produced response $i$. For each specialist response, we construct the cross-group comparator set
\begin{equation}
\mathcal{P}_d(i)=\{j:\ \mathcal{G}_j\ne\mathcal{G}_i,\ d_j=d_i=d,\ |g_i-g_j|\ge\delta_d\}.
\label{eq:comparator-set}
\end{equation}
The minimum reference gap $\delta_d$ removes ambiguous pairs. For each valid pair, a centered BPR-inspired reward~\citep{rendle2012bpr} is
\begin{equation}
r_{ij}^{(d)}=2\,\sigma_L\!\left(
\frac{\operatorname{sign}(g_i-g_j)(p_i-p_j)-m_d}{\tau_d}
\right)-1,
\label{eq:pairwise}
\end{equation}
where $\sigma_L(z)=(1+e^{-z})^{-1}$, $m_d$ is the ordering margin, and $\tau_d$ controls smoothness. The reward is positive for a correct predicted order and negative for a reversal. We aggregate pair rewards using reference-score separation as a confidence weight:
\begin{equation}
R_{\mathrm{rank},i}^{(d)}=
\begin{cases}
\frac{\sum_{j\in\mathcal{P}_d(i)}|g_i-g_j|\,r_{ij}^{(d)}}
{\sum_{j\in\mathcal{P}_d(i)}|g_i-g_j|},
& \mathcal{P}_d(i)\neq\emptyset,\\[3pt]
0, & \mathcal{P}_d(i)=\emptyset.
\end{cases}
\label{eq:rank}
\end{equation}
The reference gap selects and weights reliable comparisons rather than serving as a regression target; router records are excluded. Calibration constrains absolute scores, whereas ranking constrains cross-image order. Their combination avoids trading one objective for the other: ranking need not move all predictions away from the rubric-derived scale, and calibration alone need not determine the ordering of visually close portraits.

Semantic and format rewards enforce rubric-compatible labels and parseable schemas. Router records replace calibration with intent-category and fusion-weight rewards and disable ranking. Appendices~\ref{app:reward-details}, \ref{app:policy-optimization}, and~\ref{app:experimental-details} provide the auxiliary rewards, policy objective, and RL settings, respectively.

\section{Experiments}

\subsection{Experimental setup}

We train \method{} from Qwen3.5-VL-9B in the two-stage procedure described in Section~4. The SFT stage uses seven task-specific instruction corpora constructed with the same rubric-driven annotation protocol as \bench, and the RL stage starts from the SFT adapter and optimizes the intent-conditioned reward with Dr.GRPO~\citep{liu2024understanding}. The visual encoder and aligner are frozen in both stages, and LoRA is applied to the language backbone. Full optimization settings, reward weights, and compute details are provided in Appendix~\ref{app:experimental-details}.

\subsection{Evaluation protocol}

The standard \bench{} evaluation contains 1.6K images expanded into 11K task-specific examples, and the hard-case evaluation contains 1K held-out images expanded into 7K task examples. We report Pearson's linear correlation coefficient (PLCC) for score calibration and Spearman's rank correlation coefficient (SRCC) for ordinal consistency. Comparisons cover general-purpose MLLMs, the open InternVL3-8B checkpoint~\citep{zhu2025internvl3}, and specialized aesthetic or perceptual assessment systems, including PortraitCraft~\citep{sha2026portraitcraft,bai2025qwen3}, ArtiMuse~\citep{cao2026artimuse}, UniPercept~\citep{cao2025unipercept}, HumanAesExpert-8B~\citep{liao2025humanaesexpert}, and Q-Align~\citep{wu2023q}. \method{} overall scores are fused using predicted intent-routing weights, whereas baselines without an intent router use their native holistic outputs. For holistic-only scorers, daggered dimension entries report the correlation between their single holistic prediction and each \bench{} target as a diagnostic rather than a dimension-conditioned estimate. Additional baseline and split details are given in Appendix~\ref{app:experimental-details}.

\begin{table}[t]
\caption{Standard \bench{} results (PLCC/SRCC; higher is better). $\dagger$ denotes diagnostic correlations from a holistic-only scorer.}
\label{tab:standard-results}
\begin{center}
\tiny
\setlength{\tabcolsep}{1.7pt}
\renewcommand{\arraystretch}{0.94}
\resizebox{\linewidth}{!}{%
\begin{tabular}{l*{14}{c}}
\toprule
\multirow{2}{*}{Model} & \multicolumn{2}{c}{Structure} & \multicolumn{2}{c}{Composition}
& \multicolumn{2}{c}{Quality} & \multicolumn{2}{c}{Tone}
& \multicolumn{2}{c}{Theme} & \multicolumn{2}{c}{Emotion}
& \multicolumn{2}{c}{Overall} \\
\cmidrule(lr){2-3}\cmidrule(lr){4-5}\cmidrule(lr){6-7}\cmidrule(lr){8-9}
\cmidrule(lr){10-11}\cmidrule(lr){12-13}\cmidrule(lr){14-15}
& PLCC & SRCC & PLCC & SRCC & PLCC & SRCC & PLCC & SRCC
& PLCC & SRCC & PLCC & SRCC & PLCC & SRCC \\
\midrule
Qwen3.5-VL-9B
& \underline{0.833} & 0.822 & 0.724 & 0.742 & 0.741 & 0.698 & \underline{0.807} & 0.789
& 0.706 & 0.714 & 0.698 & \underline{0.740} & \underline{0.857} & \underline{0.874} \\
InternVL3-8B
& 0.393 & 0.453 & 0.518 & 0.505 & 0.581 & 0.623 & 0.530 & 0.570 & 0.448 & 0.511 & 0.297 & 0.296 & 0.607 & 0.668 \\
PortraitCraft-4B$^{\dagger}$
& 0.698 & 0.801 & 0.647 & 0.760 & 0.738 & 0.773 & 0.578 & 0.698
& 0.613 & 0.754 & 0.535 & 0.655 & 0.652 & 0.785 \\
Qwen3.5 Flash
& 0.791 & \underline{0.837} & 0.620 & 0.700 & \underline{0.807} & \underline{0.798} & 0.795 & \underline{0.818}
& 0.590 & 0.705 & 0.667 & 0.723 & 0.804 & 0.763 \\
ArtiMuse$^{\dagger}$
& 0.741 & 0.785 & \underline{0.744} & \underline{0.796} & 0.556 & 0.602 & 0.688 & 0.717
& \underline{0.725} & \underline{0.756} & \underline{0.712} & 0.703 & 0.785 & 0.804 \\
UniPercept$^{\dagger}$
& 0.648 & 0.709 & 0.665 & 0.711 & 0.624 & 0.648 & 0.631 & 0.655
& 0.676 & 0.711 & 0.625 & 0.652 & 0.703 & 0.729 \\
HumanAesExpert-8B
& 0.537 & 0.574 & 0.541 & 0.567 & 0.583 & 0.546 & 0.465 & 0.483
& 0.414 & 0.460 & 0.361 & 0.402 & 0.457 & 0.499 \\
Q-Align$^{\dagger}$
& 0.717 & 0.764 & 0.729 & 0.752 & 0.715 & 0.705 & 0.679 & 0.697
& 0.652 & 0.684 & 0.604 & 0.622 & 0.707 & 0.724 \\
\midrule
\method{} (Ours)
& \textbf{0.875} & \textbf{0.912} & \textbf{0.878} & \textbf{0.903}
& \textbf{0.889} & \textbf{0.891} & \textbf{0.914} & \textbf{0.913}
& \textbf{0.879} & \textbf{0.901} & \textbf{0.847} & \textbf{0.863}
& \textbf{0.924} & \textbf{0.934} \\
\bottomrule
\end{tabular}%
}
\end{center}
\end{table}

\begin{table}[t]
\caption{\bench{} hard-case results (PLCC/SRCC; higher is better). $\dagger$ denotes diagnostic correlations from a holistic-only scorer.}
\label{tab:hard-results}
\begin{center}
\tiny
\setlength{\tabcolsep}{1.7pt}
\renewcommand{\arraystretch}{0.94}
\resizebox{\linewidth}{!}{%
\begin{tabular}{l*{14}{c}}
\toprule
\multirow{2}{*}{Model} & \multicolumn{2}{c}{Structure} & \multicolumn{2}{c}{Composition}
& \multicolumn{2}{c}{Quality} & \multicolumn{2}{c}{Tone}
& \multicolumn{2}{c}{Theme} & \multicolumn{2}{c}{Emotion}
& \multicolumn{2}{c}{Overall} \\
\cmidrule(lr){2-3}\cmidrule(lr){4-5}\cmidrule(lr){6-7}\cmidrule(lr){8-9}
\cmidrule(lr){10-11}\cmidrule(lr){12-13}\cmidrule(lr){14-15}
& PLCC & SRCC & PLCC & SRCC & PLCC & SRCC & PLCC & SRCC
& PLCC & SRCC & PLCC & SRCC & PLCC & SRCC \\
\midrule
Qwen3.5-VL-9B
& 0.515 & 0.513 & 0.346 & 0.281 & 0.558 & 0.506 & 0.602 & 0.574
& 0.385 & \underline{0.474} & 0.391 & \underline{0.413} & \underline{0.649} & 0.629 \\
InternVL3-8B
& 0.030 & 0.036 & 0.156 & 0.112 & 0.402 & 0.476 & 0.228 & 0.273
& -0.010 & -0.033 & 0.121 & 0.184 & 0.210 & 0.215 \\
PortraitCraft-4B$^{\dagger}$
& 0.406 & 0.403 & 0.361 & 0.331 & 0.479 & 0.463 & 0.151 & 0.145
& 0.342 & 0.328 & 0.288 & 0.297 & 0.406 & 0.400 \\
Qwen3.5 Flash
& \underline{0.522} & \underline{0.522} & 0.240 & 0.214 & \underline{0.578} & \underline{0.543} & \underline{0.624} & \underline{0.601}
& 0.304 & 0.411 & \underline{0.407} & 0.400 & 0.615 & \underline{0.635} \\
ArtiMuse$^{\dagger}$
& 0.430 & 0.421 & \underline{0.407} & \underline{0.398} & 0.336 & 0.346 & 0.368 & 0.350
& \underline{0.410} & 0.418 & 0.387 & 0.383 & 0.501 & 0.495 \\
UniPercept$^{\dagger}$
& 0.292 & 0.319 & 0.296 & 0.271 & 0.322 & 0.349 & 0.151 & 0.146
& 0.322 & 0.331 & 0.296 & 0.294 & 0.352 & 0.358 \\
HumanAesExpert-8B
& 0.243 & 0.237 & 0.197 & 0.194 & 0.200 & 0.178 & 0.110 & 0.105
& 0.106 & 0.077 & 0.100 & 0.094 & 0.127 & 0.154 \\
Q-Align$^{\dagger}$
& 0.422 & 0.400 & 0.360 & 0.339 & 0.451 & 0.411 & 0.223 & 0.215
& 0.277 & 0.242 & 0.229 & 0.227 & 0.374 & 0.348 \\
\midrule
\method{} (Ours)
& \textbf{0.679} & \textbf{0.650} & \textbf{0.646} & \textbf{0.614}
& \textbf{0.768} & \textbf{0.750} & \textbf{0.789} & \textbf{0.752}
& \textbf{0.689} & \textbf{0.627} & \textbf{0.650} & \textbf{0.629}
& \textbf{0.829} & \textbf{0.795} \\
\bottomrule
\end{tabular}%
}
\end{center}
\end{table}

\subsection{Benchmark results and model analysis}

\method{} leads every standard-set metric, reaching mean specialist PLCC/SRCC values of 0.880/0.897 and overall values of 0.924/0.934 (\autoref{tab:standard-results}). Its largest gains are on theme and style and emotion, which require intent-aware judgments of pose, scene, styling, and affect.

General-purpose MLLMs remain uneven, while holistic systems such as ArtiMuse, UniPercept, and Q-Align provide only diagnostic covariance with dimension targets. PortraitCraft-4B transfers better to structure and composition but is weaker on tone, theme and style, and emotion. By predicting intent and specialist scores before fusion, \method{} makes the final estimate traceable to benchmark criteria.

\subsection{Hard-case analysis}

The hard-case set narrows the overall-score standard deviation from 0.688 to 0.315 and the mean specialist deviation from 0.913 to 0.457 (panel~(b) of \autoref{fig:overall-score-distributions}). In this compressed range, \method{} retains mean specialist PLCC/SRCC values of 0.704/0.670 and overall values of 0.829/0.795 (\autoref{tab:hard-results}). Its margin over ArtiMuse~\citep{cao2026artimuse} grows from 0.139/0.130 on the standard benchmark to 0.328/0.300 on the hard set, highlighting the strength of calibrated ranking on visually similar portraits.

Cross-group BPR is especially helpful for structure and composition, where small framing or subject-presentation differences can reverse order without producing large score gaps. Gaussian calibration anchors the score scale, whereas ranking directly penalizes same-dimension ordinal inversions (\autoref{tab:hard-ablation}).
\subsection{Component ablations}
\label{sec:component-ablations}

The ablations isolate the effect of structured SFT, Gaussian calibration, and cross-group BPR before intent-conditioned fusion.

\begin{table}[htbp]
\caption{Component ablation on the standard \bench. Each result is reported as PLCC/SRCC; higher is better. Checkmarks denote enabled training components.}
\label{tab:standard-ablation}
\begin{center}
\scriptsize
\setlength{\tabcolsep}{2.8pt}
\renewcommand{\arraystretch}{1.06}
\resizebox{\linewidth}{!}{%
\begin{tabular}{lccc*{7}{c}}
\toprule
\multirow{2}{*}{Method} & \multicolumn{3}{c}{Training component}
& \multicolumn{7}{c}{PLCC/SRCC} \\
\cmidrule(lr){2-4}\cmidrule(lr){5-11}
& SFT & Gaussian & Cross-BPR
& Structure & Composition & Quality & Tone & Theme & Emotion & Overall \\
\midrule
Qwen3.5-VL-9B
& & & & 0.833/0.822 & 0.724/0.742 & 0.741/0.698 & 0.807/0.789
& 0.706/0.714 & 0.698/0.740 & 0.857/0.874 \\
\method{}-SFT
& $\checkmark$ & & & 0.856/0.896 & 0.852/0.885 & \underline{0.873}/\underline{0.868} & 0.904/0.905
& \underline{0.855}/0.877 & 0.810/0.830 & 0.909/0.917 \\
\method{}-GRPO (Gaussian only)
& $\checkmark$ & $\checkmark$ & & \underline{0.863}/\underline{0.901} & \underline{0.869}/\underline{0.897} & 0.872/\underline{0.868} & \underline{0.907}/\underline{0.907}
& 0.854/\underline{0.883} & \underline{0.820}/\underline{0.837} & \underline{0.915}/\underline{0.924} \\
\method{}-GRPO
& $\checkmark$ & $\checkmark$ & $\checkmark$
& \textbf{0.875/0.912} & \textbf{0.878/0.903} & \textbf{0.889/0.891}
& \textbf{0.914/0.913} & \textbf{0.879/0.901} & \textbf{0.847/0.863}
& \textbf{0.924/0.934} \\
\bottomrule
\end{tabular}%
}
\end{center}
\end{table}

\begin{table}[htbp]
\caption{Component ablation on the \bench{} hard-case set. Each result is reported as PLCC/SRCC; higher is better. Checkmarks denote enabled training components.}
\label{tab:hard-ablation}
\begin{center}
\scriptsize
\setlength{\tabcolsep}{2.8pt}
\renewcommand{\arraystretch}{1.06}
\resizebox{\linewidth}{!}{%
\begin{tabular}{lccc*{7}{c}}
\toprule
\multirow{2}{*}{Method} & \multicolumn{3}{c}{Training component}
& \multicolumn{7}{c}{PLCC/SRCC} \\
\cmidrule(lr){2-4}\cmidrule(lr){5-11}
& SFT & Gaussian & Cross-BPR
& Structure & Composition & Quality & Tone & Theme & Emotion & Overall \\
\midrule
Qwen3.5-VL-9B
& & & & 0.515/0.513 & 0.346/0.281 & 0.558/0.506 & 0.602/0.574
& 0.385/0.474 & 0.391/0.413 & 0.649/0.629 \\
\method{}-SFT
& $\checkmark$ & & & 0.562/\textbf{0.668} & 0.558/0.524 & \underline{0.737}/\underline{0.723} & \textbf{0.795}/\textbf{0.753}
& 0.570/\underline{0.598} & 0.549/0.559 & 0.780/\underline{0.758} \\
\method{}-GRPO (Gaussian only)
& $\checkmark$ & $\checkmark$ & & \underline{0.617}/0.624 & \underline{0.597}/\underline{0.540} & 0.713/0.694 & 0.768/0.731
& \underline{0.661}/0.590 & \underline{0.610}/\underline{0.587} & \underline{0.789}/0.755 \\
\method{}-GRPO
& $\checkmark$ & $\checkmark$ & $\checkmark$
& \textbf{0.679}/\underline{0.650} & \textbf{0.646}/\textbf{0.614} & \textbf{0.768}/\textbf{0.750}
& \underline{0.789}/\underline{0.752} & \textbf{0.689}/\textbf{0.627} & \textbf{0.650}/\textbf{0.629}
& \textbf{0.829}/\textbf{0.795} \\
\bottomrule
\end{tabular}%
}
\end{center}
\end{table}

On the standard set, structured SFT raises overall PLCC/SRCC from 0.857/0.874 to 0.909/0.917, Gaussian calibration reaches 0.915/0.924, and cross-group BPR reaches 0.924/0.934 (\autoref{tab:standard-ablation}). On the hard set, BPR raises the Gaussian-only result from 0.789/0.755 to 0.829/0.795 and improves all six dimensions. Calibration therefore stabilizes absolute scores, while cross-group ranking preserves fine-grained order in a compressed score range.

\section{Conclusion}

We introduced \bench, an 11K-scale benchmark that represents portrait aesthetics as intent-conditioned judgments across nine photographic intents and six dimensions. Its router--specialist--fusion workflow turns expert rubrics into structured supervision with scores, labels, visual evidence, and rule decisions. \method{} learns the same task structure through multi-task SFT, then combines Gaussian calibration with same-dimension cross-image ranking. Results on the standard and hard-case sets show that this combination improves both agreement with rubric-derived scores and ordinal consistency over the evaluated general-purpose and specialized baselines.


\clearpage




\newpage
\bibliography{iclr2027}
\bibliographystyle{iclr2027_conference}

\clearpage
\appendix
\section*{APPENDIX}

This appendix provides compact supporting material for interpreting the benchmark scope, diagnostic figures, ablation results, and implementation settings.

\section{Resource Comparison}
\label{app:resource-comparison}

Table~\ref{tab:resource-comparison} uses five coarse properties (scale, portrait focus, supervision type, dimensional structure, and textual annotation) to clarify how the benchmark differs from prior resources.

\begin{table}[htbp]
\caption{Comparison with representative aesthetics and portrait-related resources. \bench{} emphasizes intent-conditioned, multi-dimensional portrait understanding rather than a single global score.}
\label{tab:resource-comparison}
\begin{center}
\small
\resizebox{\linewidth}{!}{%
\begin{tabular}{llllll}
\toprule
Resource & Scale & Port. & Supervision & Dim. & Text \\
\midrule
AVA~\citep{murray2012ava} & 255K & no & ratings & global & limited \\
AADB~\citep{kong2016photo} & 10K & no & score/attr./rank & attr. & no \\
ArtiMuse~\citep{cao2026artimuse} & 10K & no & score/attr./text & 8 attr. & yes \\
UniPercept~\citep{cao2025unipercept} & 800K & no & public VR/VQA & perception & yes \\
Qwen-Image-Bench~\citep{li2026qwen} & 1K & no & judge scores & 5D/56F & yes \\
MPAD/MPAAD~\citep{huang2022aesthetic} & 18K/3K & yes & overall/attr. & 3 attr. & no \\
PIQ23~\citep{chahine2023image} & 5K & yes & pairwise & quality & no \\
Q-Bench-Portrait~\citep{wu2026q} & 2.8K & yes & QA & quality & yes \\
PortraitCraft~\citep{sha2026portraitcraft} & 50K & yes & score/attr./VQA & comp. & yes \\
\midrule
\bench{} & 11K & yes & intent/6 scores/labels & 6D & yes \\
\bottomrule
\end{tabular}%
}
\end{center}
\end{table}

\section{Diagnostic Visualizations}
\label{app:diagnostics}

This diagnostic reports how score mass varies across specialist dimensions.


\begin{figure}[H]
\begin{center}
\includegraphics[width=\linewidth]{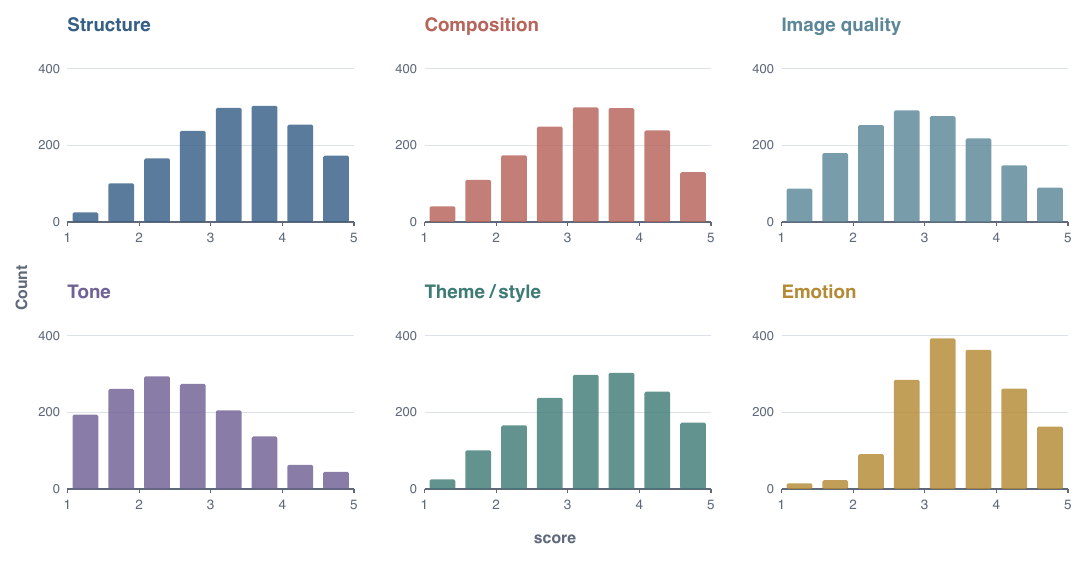}
\end{center}
\caption{Specialist-score distributions in \bench. The six portrait-aesthetic dimensions show distinct score profiles rather than a single shared difficulty pattern, supporting dimension-wise evaluation of structure, composition, image quality, tone, theme/style, and emotion.}
\label{fig:dimension-score-distributions}
\end{figure}

\section{Policy Optimization Details}
\label{app:policy-optimization}

For an image--instruction pair $x=(I,q_a)$, let $y_i\sim\pi_{\theta_{\mathrm{old}}}(\cdot\mid x)$ be the $i$-th of $K$ sampled structured responses. Following group relative policy optimization (GRPO)~\citep{shao2024deepseekmath}, we center rewards within each prompt group and scale them by the dispersion of the rollout batch,
\begin{equation}
\widehat{A}_i=\frac{R_i-\mu_{\mathcal{G}(x)}}{s_{\mathcal{B}}+\epsilon_A},
\label{eq:advantage}
\end{equation}
where $\mathcal{G}(x)$ is the completion group for $x$ and $\mathcal{B}$ is the rollout batch. The policy is optimized with a clipped surrogate objective regularized toward the SFT reference policy $\pi_{\mathrm{ref}}$:
\begin{equation}
\begin{split}
\mathcal{J}(\theta)=\mathbb{E}\!\left[\frac{1}{K}\sum_{i=1}^{K}\frac{1}{Z}
\sum_{t=1}^{|y_i|}\left(
\min\!\left(\rho_{i,t}\widehat{A}_i,
\operatorname{clip}(\rho_{i,t},1-\epsilon,1+\epsilon)\widehat{A}_i\right)
-\beta_{\mathrm{KL}}D_{\mathrm{KL}}\!\left(\pi_{\theta}\,\|\,\pi_{\mathrm{ref}}\right)_{i,t}
\right)\right],
\end{split}
\label{eq:grpo-objective}
\end{equation}
where $\rho_{i,t}=\pi_{\theta}(y_{i,t}\mid x,y_{i,<t})/\pi_{\theta_{\mathrm{old}}}(y_{i,t}\mid x,y_{i,<t})$, $\epsilon$ is the clipping radius, and $Z$ is a fixed response-length normalizer. The clipped ratio limits destructive policy updates, while the token-level KL term controls drift from the schema-aware SFT model.

\section{Auxiliary Reward Definitions}
\label{app:reward-details}

\textbf{Intent routing ($R_{\mathrm{route}}$).}
The router has no scalar aesthetic target. Its reward instead combines intent-category accuracy with the closeness of the predicted fusion-weight vector $\widehat{\mathbf{w}}$ to its reference $\mathbf{w}$:
\begin{equation}
R_{\mathrm{route}}
=\eta_r\mathbf{1}[\widehat{c}=c]
+(1-\eta_r)\left[1-\tfrac{1}{2}\|\widehat{\mathbf{w}}-\mathbf{w}\|_1\right]_{+}.
\label{eq:router-reward}
\end{equation}

\textbf{Semantic consistency ($R_{\mathrm{sem}}$).}
Numeric agreement is insufficient when a response assigns an incompatible semantic label. Let $s_d(y_i)$ and $s_d^\star$ be the task-specific semantic field extracted from the response and the reference. We use a compact task-specific matcher:
\begin{equation}
R_{\mathrm{sem}}(y_i,d)=\rho_d\!\left(s_d(y_i),s_d^\star\right),
\label{eq:semantic-reward}
\end{equation}
where $\rho_d$ assigns full credit to exact semantic matches, partial credit to compatible labels, a penalty to contradictions, a parse-failure penalty to malformed structured responses, and a neutral value when no semantic target is defined. For example, in theme and style, two subject codes with the same major code, such as $S3.1$ and $S3.4$, are compatible and receive partial credit; emotion uses nonempty category-token overlap as the compatibility check.
\par
{\bfseries Structural validity ($R_{\mathrm{fmt}}$).}
The format reward grades dimension-specific schema completeness and subtracts a length-dependent penalty,
\begin{equation}
R_{\mathrm{fmt}}(y,d)=\operatorname{clip}\!\left(B_d(y)-\ell_d(|y|),\,r_{\min},r_{\max}\right),
\label{eq:format-reward}
\end{equation}
where $B_d$ checks well-formed structured responses and required fields, and $\ell_d$ discourages unnecessarily long responses. Malformed outputs receive a parse-failure penalty and are excluded from ordinal comparison.

\clearpage
\section{Additional Experimental Details}
\label{app:experimental-details}
\paragraph{Training setup.}
\method{} is initialized from Qwen3.5-VL-9B. For supervised fine-tuning, we apply LoRA with rank~64 and scaling factor~128 to all linear modules while freezing the visual encoder and aligner. SFT is performed in bfloat16 for three epochs using a cosine learning-rate schedule with a peak learning rate of $4\times10^{-5}$ and 5\% warmup. The maximum sequence length is 16,384, the per-device batch size is~1, and gradient accumulation is~2. Gradient checkpointing and ZeRO-3 offload are enabled, and training uses 24 NVIDIA L20Z GPUs.

For reinforcement learning, the corpus contains approximately 2,657 images expanded into 18,600 task-specific examples. Both the trainable policy and reference policy are initialized from the seven-task SFT adapter, with the visual encoder and aligner remaining frozen. We optimize the intent-conditioned reward in Section~4.3 using Dr.GRPO with batch-level reward scaling. Each prompt samples eight completions, and training runs for one epoch with a constant learning rate of $2\times10^{-6}$ and KL coefficient~$0.02$, using bfloat16 precision and ZeRO-3 across 24 NVIDIA L20Z GPUs.

\paragraph{Evaluation details.}
The standard evaluation uses one routing task and six specialist rating tasks for each of 1,600 images. The hard-case set is image-disjoint and contains 1,000 images from a held-out portrait pool after exact- and basename-level exclusion against training and earlier evaluation artifacts. General-purpose MLLMs are evaluated with the corresponding \bench{} rubric prompt; Qwen3.5 Flash is queried through Alibaba Cloud Model Studio using the fixed API identifier \texttt{qwen3.5-flash}~\citep{qwen35flash2026}. Specialized aesthetic models are evaluated with their released checkpoints and native scoring interfaces. All table entries are PLCC/SRCC, and unavailable evaluations are marked with ``--''.

\paragraph{Intent-conditioned fusion.}
Table~\ref{tab:intent-weights} reports the router weights used to convert six specialist scores into an overall portrait-aesthetic score. Functional and commercial portraits emphasize quality and presentation, aesthetic and social portraits rely more on tone and theme and style, and snapshot or symbolic portraits shift toward emotion and meaning.
\begin{table}[htbp]
\caption{Intent-conditioned fusion weights used by the router.}
\label{tab:intent-weights}
\centering
\scriptsize
\setlength{\tabcolsep}{3.6pt}
\renewcommand{\arraystretch}{0.82}
\begin{tabular}{lrrrrrr}
\toprule
Intent & Struct. & Comp. & Quality & Tone & Theme & Emotion \\
\midrule
I Functional & 0.20 & 0.20 & 0.30 & 0.15 & 0.10 & 0.05 \\
II Commercial & 0.10 & 0.15 & 0.35 & 0.25 & 0.10 & 0.05 \\
III Narrative & 0.15 & 0.15 & 0.15 & 0.20 & 0.10 & 0.25 \\
IV Aesthetic & 0.10 & 0.15 & 0.15 & 0.30 & 0.20 & 0.10 \\
V Social & 0.05 & 0.15 & 0.15 & 0.20 & 0.30 & 0.15 \\
VI Artistic & 0.10 & 0.15 & 0.05 & 0.20 & 0.20 & 0.30 \\
VII Hybrid & 0.10 & 0.15 & 0.15 & 0.20 & 0.20 & 0.20 \\
VIII Snapshot & 0.05 & 0.10 & 0.05 & 0.15 & 0.15 & 0.50 \\
IX Symbolic & 0.00 & 0.05 & 0.00 & 0.05 & 0.30 & 0.60 \\
\bottomrule
\end{tabular}
\end{table}
\paragraph{Rank-aligned GRPO.}
Table~\ref{tab:grpo-hyperparameters} lists the RL-stage settings. Calibration and cross-image ranking dominate the reward, while semantic and format terms remain lightweight constraints.
\begin{table}[htbp]
\caption{Hyperparameters for rank-aligned GRPO.}
\label{tab:grpo-hyperparameters}
\centering
\scriptsize
\setlength{\tabcolsep}{5pt}
\renewcommand{\arraystretch}{0.82}
\resizebox{\linewidth}{!}{%
\begin{tabular}{ll@{\hspace{18pt}}ll}
\toprule
Hyperparameter & Value & Hyperparameter & Value \\
\midrule
Completions per prompt $K$ & 8 & Training epochs & 1 \\
Learning rate & $2\times10^{-6}$ & Warmup ratio & 0.05 \\
KL coefficient & 0.02 & Clip radius & 0.12 \\
Per-device batch size & 4 & Gradient accumulation & 1 \\
Calibration/ranking weights & 1.4 / 0.8 & Semantic/format weights & 0.05 / 0.02 \\
Gaussian bandwidth $\sigma$ & 0.5 & Primary/sub-score weights & 0.8 / 0.2 \\
Minimum reference gap $\delta$ & 1.0 & BPR margin / temperature & 0.0 / 0.5 \\
Sampling temperature / top-$p$ & 1.0 / 0.95 & Maximum lengths & 512 / 16,384 \\
Maximum gradient norm & 0.5 & Precision & bfloat16 \\
\bottomrule
\end{tabular}%
}
\end{table}

\clearpage
\section{Examples}
\label{app:qualitative-predictions}

\begin{figure}[H]
\centering
\includegraphics[width=\linewidth]{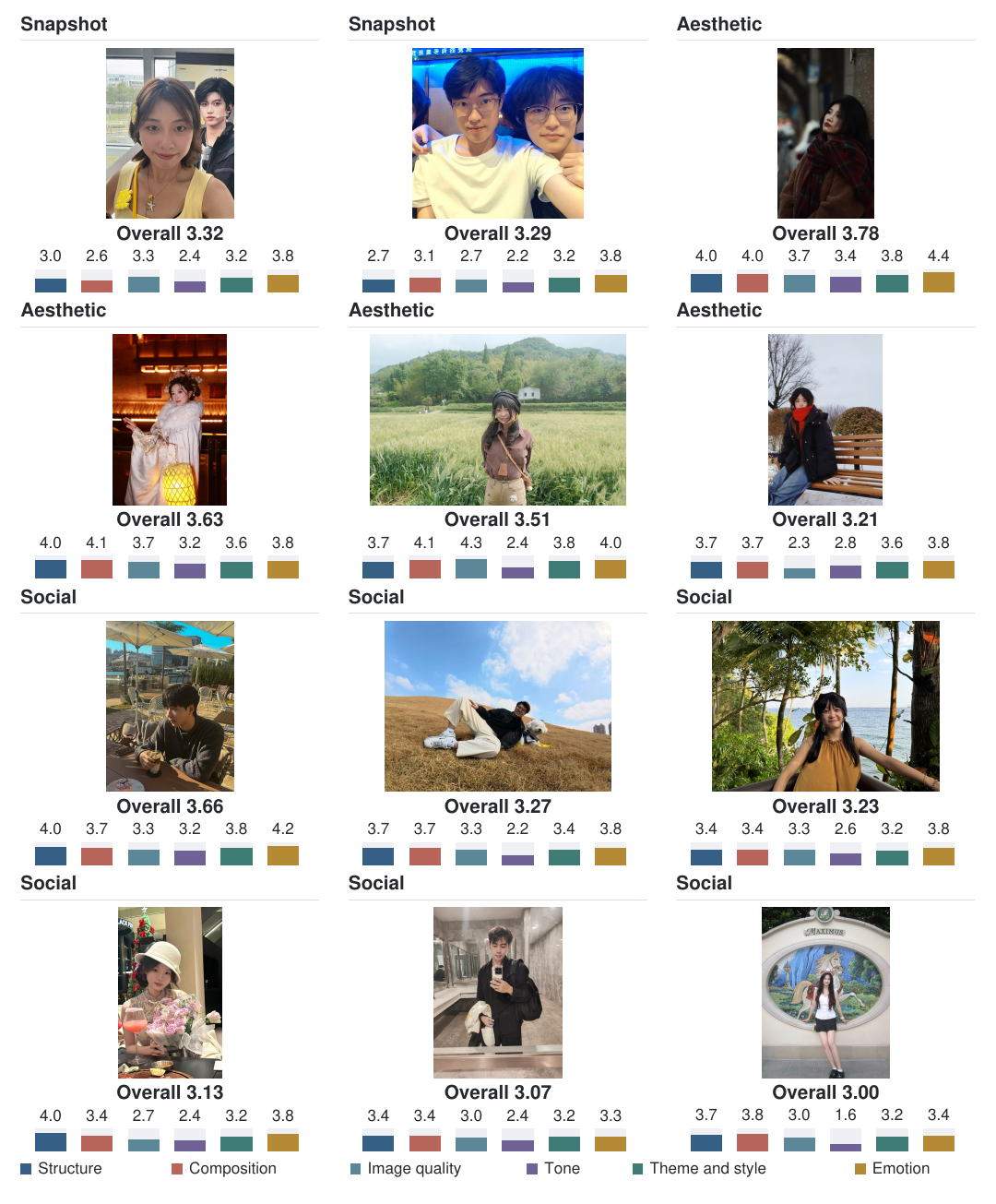}
\caption{Structured predictions for twelve portraits across six intent categories, grouped by category and ordered by decreasing overall score. Colored bars show the six specialist scores.}
\label{fig:qualitative-score-cases-a}
\end{figure}

\clearpage


 
\begin{figure}[H]
\centering
\includegraphics[width=\linewidth]{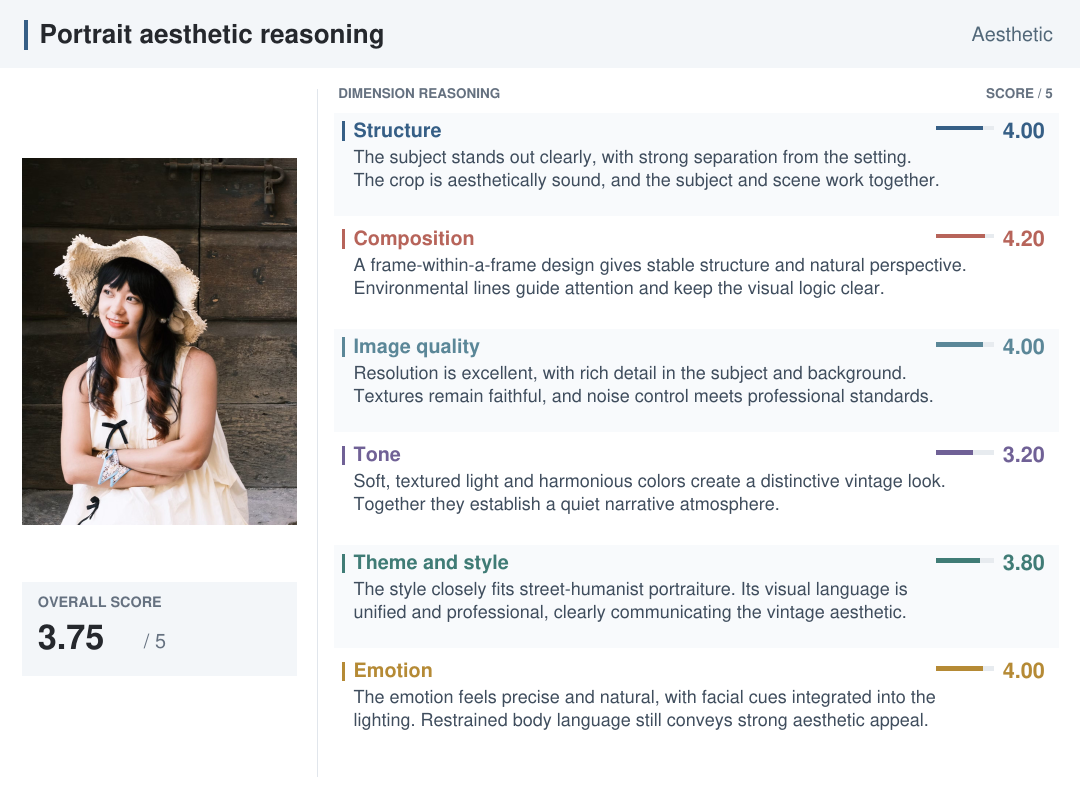}
\caption{Dimension-level aesthetic explanations for an aesthetic portrait with an overall score of 3.75.}
\label{fig:reasoning-aesthetic}
\end{figure}

\clearpage

\begin{figure}[H]
\centering
\includegraphics[width=\linewidth]{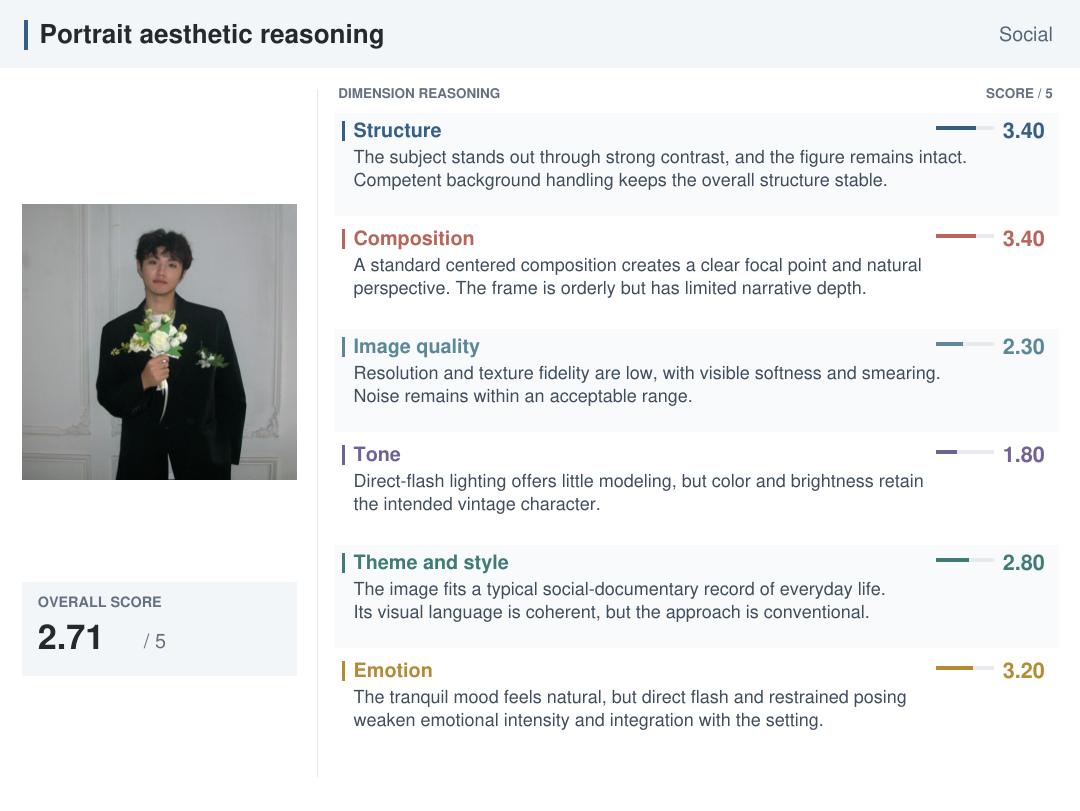}
\caption{Dimension-level aesthetic explanations for a social portrait with an overall score of 2.71.}
\label{fig:reasoning-snapshot}
\end{figure}




\end{document}